\documentclass[10pt,twocolumn,letterpaper]{article}

\usepackage{iccv}
\usepackage{times}
\usepackage{epsfig}
\usepackage{graphicx}
\usepackage{subfigure}
\usepackage{amsmath}
\usepackage{amssymb}
\usepackage{multirow}
\usepackage{color}
\usepackage{epstopdf}
\usepackage{caption}

\iccvfinalcopy 

\ificcvfinal\fi
\begin{document}

\title{Tube Convolutional Neural Network (T-CNN) for Action Detection in Videos}

\author{Rui Hou, Chen Chen, Mubarak Shah\\
Center for Research in Computer Vision (CRCV), University of Central Florida (UCF)\\
{\tt\small houray@gmail.com, chenchen870713@gmail.com, shah@crcv.ucf.edu}
}

\maketitle

\begin{abstract}
Deep learning has been demonstrated to achieve excellent results for image classification and object detection.
However, the impact of deep learning on video analysis (\eg action detection and recognition) has been limited due to complexity of video data and lack of annotations. 
Previous convolutional neural networks (CNN) based video action detection approaches usually consist of two major steps: frame-level action proposal generation and association of proposals across frames. Also, most of these methods employ two-stream CNN framework to handle spatial and temporal feature separately.  In this paper, we propose an end-to-end deep network called Tube Convolutional Neural Network (T-CNN) for action detection in videos. The proposed architecture is a unified deep network that is able to recognize and localize action based on 3D convolution features.
A video is first divided into equal length clips and next for each clip a set of tube proposals are generated  based on 3D Convolutional Network (ConvNet) features. 
Finally, the tube proposals of different clips are linked together employing network flow and spatio-temporal action detection is performed using these linked video proposals.
Extensive experiments on several video datasets demonstrate the superior performance  of T-CNN for classifying and localizing actions in both trimmed and untrimmed videos compared to state-of-the-arts.

\end{abstract}
	
\section{Introduction}
The goal of action detection is to detect every occurrence of a given action within a long video, and to localize each detection both in space and time. Deep learning learning based approaches have significantly improved video action recognition performance. Compared to action {\em recognition}, action {\em detection} is a more challenging task due to flexible volume shape and large spatio-temporal search space.

\begin{figure}[t]
	\centering
	\includegraphics[width=0.65\linewidth]{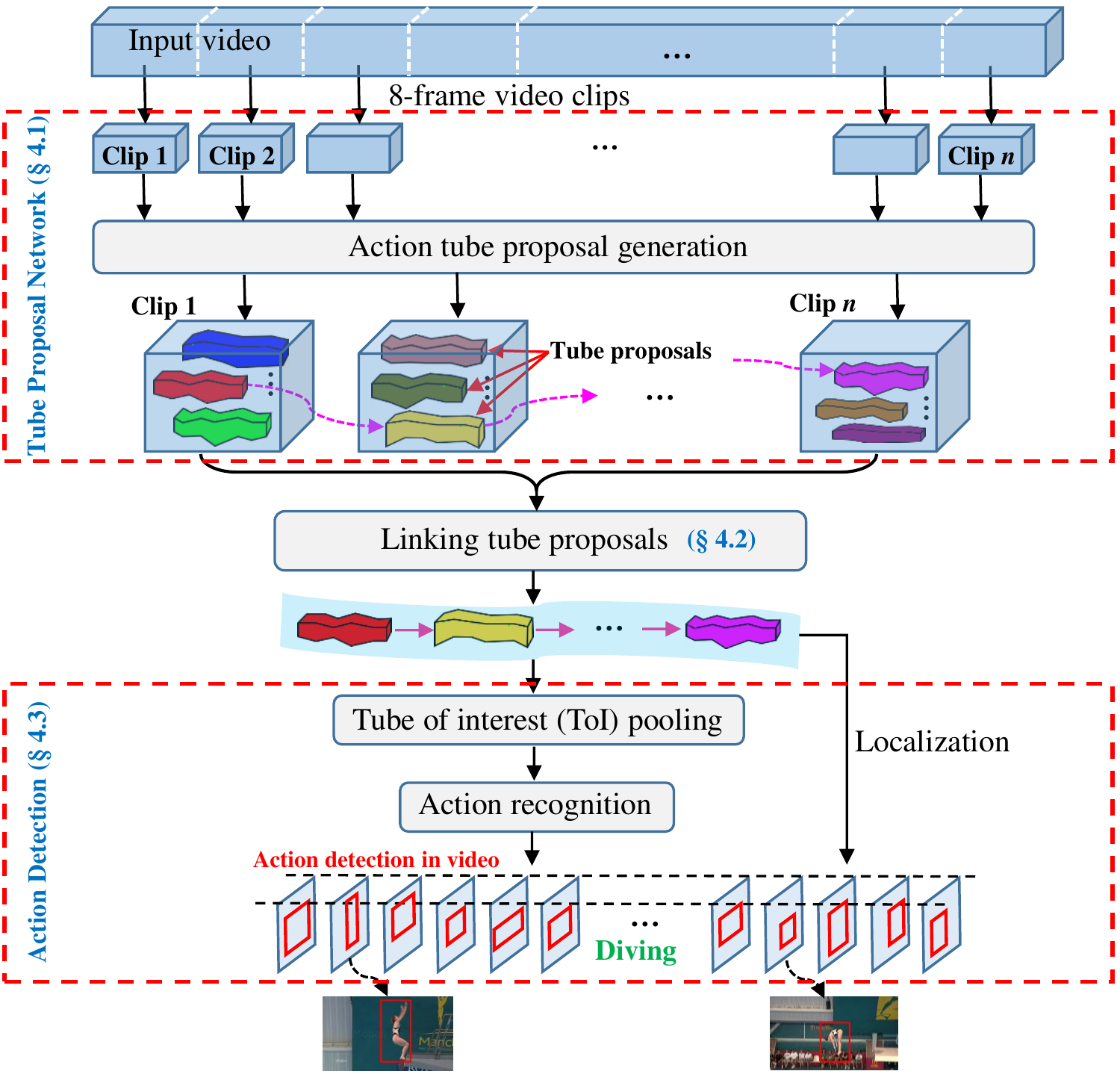}
	\caption{Overview of the proposed Tube Convolutional Neural Network (T-CNN).}
	\label{fig:Teaser}
\end{figure}

Previous deep learning based action detection approaches first detect frame-level action proposals by popular proposal algorithms \cite{gkioxari2015finding,weinzaepfel2015learning} or by training proposal networks \cite{peng2016multi}. Then the frame-level action proposals are associated across frames to form final action detections through tracking based approaches. Moreover, in order to capture both spatial and temporal information of an action, two-stream networks (a spatial CNN and a motion CNN) are used. In this manner, the spatial and motion information are processed separately.

Region Convolution Neural Network (R-CNN) for object detection in images was proposed by Girshick \etal \cite{rcnn_Girshick_2014_CVPR}. It was followed by a fast R-CNN proposed in \cite{fast_rcnn_Girshick_2015_ICCV}, which includes the classifier as well. Later, faster R-CNN \cite{faster_rcnn} was developed by introducing a region proposal network. It has been extensively used to produce excellent results for object detection in images.
 A natural generalization of the R-CNN from 2D images to 3D spatio-temporal volumes is to study their effectiveness for the problem of action detection in videos. A straightforward spatio-temporal generalization of the R-CNN approach would be to treat action detection in videos as a set of 2D image detections using faster R-CNN. However, unfortunately, this approach does not take the temporal information into account and is not sufficiently expressive to distinguish between actions.

Inspired by the pioneering work of faster R-CNN, we propose Tube Convolutional Neural Network (T-CNN) for action detection. To better capture the spatio-temporal information of video, we exploit 3D ConvNet for action detection, since it is able to capture motion characteristics in videos and shows promising result on video action recognition. We propose a novel framework by leveraging the descriptive power of 3D ConvNet. In our approach, an input video is divided into equal length clips first. Then, the clips are fed into Tube Proposal Network (TPN) and a set of tube proposals are obtained. Next, tube proposals from each video clip are linked according to their actionness scores and overlap between adjacent proposals to form a complete tube proposal for spatio-temporal action localization in the video. Finally, the Tube-of-Interest (ToI) pooling is applied to the linked action tube proposal to generate a fixed length feature vector for action label prediction.

Our work makes the following contributions:




$\bullet$ We propose an end-to-end deep learning based approach for action detection in videos. It directly operates on the original videos and captures spatio-temporal information using a single 3D network to perform action localization and recognition based on 3D convolution features. To the best of our knowledge, it is the first work to exploit 3D ConvNet for action detection.

$\bullet$ We introduce a Tube Proposal Network, which leverages skip pooling in temporal domain to preserve temporal information for action localization in 3D volumes.

$\bullet$ We propose a new pooling layer -- {\bf Tube-of-Interest} (ToI) pooling layer in T-CNN. The ToI pooling layer is a 3D generalization of Region-of-Interest (RoI) pooling layer of R-CNN. It effectively alleviates the problem with variable spatial and temporal sizes of tube proposals. We show that ToI pooling can greatly improve the recognition results.

$\bullet$ We extensively evaluate our T-CNN for action detection in both trimmed videos from UCF-Sports, J-HMDB and UCF-101 datasets and untrimmed videos from THUMOS'14 dataset and achieve state-of-the-art performance. \textit{The source code of T-CNN will be released.}


\section{Related Work}\label{sec:related_work}
Convolutional Neural Networks (CNN) have been demonstrated to achieve excellent results for action recognition \cite{lecun2015deep, action_survey}. Karpathy \etal \cite{karpathy2014large}  explore various frame-level fusion methods over time. Ng \etal \cite{lstm_ng} use recurrent neural network employing the CNN feature. Since these approaches only use frame based CNN features, the temporal information is neglected. Simonyan \etal \cite{2stream_cnn_simonyan_2014two} propose the two-stream CNN approach for action recognition. Besides a classic CNN which takes images as an input, it has a separate network for optical flow. Moreover, Wang \etal fuse the trajectories and CNN features. Although these methods, which take hand-crafted temporal feature as a separate stream, show promising performance on action recognition, however, they do not employ end to end deep network and require separate computation of  optical flow  and optimization of the parameters. 3D CNN is a logical solution to this issue. Ji \etal \cite{ji20133d} propose a 3D CNN based human detector and head tracker to segment human subjects in videos. Tran \etal \cite{c3d} leverage 3D CNN for large scale action recognition problem. Sun \etal \cite{sun2015human} propose a factorization of 3D CNN and exploit multiple ways to decompose convolutional kernels. However, to the best of our knowledge, we are the first ones to exploit 3D CNN for action {\em detection}.

Compared to action recognition, action detection is a more challenging problem \cite{gaidon2013temporal,jain201515}, which has been an active area of research.
Ke \etal \cite{ke2007event} present an approach for event detection in crowded videos. Tian \etal \cite{tian2013spatiotemporal} develop Spatio-temporal Deformable Parts Model \cite{dpm} to detect actions in videos. Jain \etal \cite{jain2014action} and Soomro \etal \cite{soomro2015action} use supervoxel and selective search to localize the action boundaries. Recently, researchers have leveraged the power of deep learning for action detection. Authors in \cite{gkioxari2015finding} extract frame-level action proposals using selective search and link them using Viterbi algorithm. While in \cite{weinzaepfel2015learning}  frame-level action proposals are obtained by EdgeBox and linked by a tracking algorithm. Two-stream R-CNNs for action detection is proposed in  \cite{peng2016multi}, where a spatial Region Proposal Network (RPN) and a motion RPN are used to generate frame-level action proposals. However, these deep learning based approaches detect actions by linking frame-level action proposals and treat the spatial  and temporal features of a video separately by training two-stream CNN. Therefore, the temporal consistency in videos is not well explored in the network. In contrast, we determine action tube proposals directly from input videos and extract compact and more effective spatio-temporal features using 3D CNN.

 For object detection in images, Girshick \etal propose Region CNN (R-CNN) \cite{rcnn_Girshick_2014_CVPR}.  In their approach region proposals are extracted using selective search. Then the candidate regions are warped to a fixed size and fed into ConvNet to extract CNN features. Finally, SVM model is trained for object classification. A fast version of R-CNN, Fast R-CNN, is presented in \cite{fast_rcnn_Girshick_2015_ICCV}. Compared to the multi-stage pipeline of R-CNN, fast R-CNN incorporates object classifier in the network and trains object classifier and bounding box regressor simultaneously. Region of interest (RoI) pooling layer is introduced to extract fixed-length feature vectors for bounding boxes with different sizes. Recently, faster R-CNN is proposed in \cite{faster_rcnn}. It introduces a RPN (Region Proposal Network) to replace selective search for proposal generation. RPN shares full image convolutional features with the detection network, thus the proposal generation is almost cost-free. Faster R-CNN achieves state-of-the-art object detection performance while being efficient during testing. Motivated by its high performance, in this paper we explore generalizing faster R-CNN from 2D image regions to 3D video volumes for action detection.


\section{Generalizing R-CNN from 2D to 3D}
\label{sec:difference}
Generalizing R-CNN from 2D image regions to 3D video tubes is challenging due to the asymmetry between space and time.
Different from images which can be cropped and reshaped into a fixed size, videos vary widely in temporal dimension. Therefore, we divide input videos into fixed length (8 frames) clips, so that video clips  can be processed with a fixed-size ConvNet architecture. Also, clip based processing mitigates the cost of GPU memory.

\begin{table}
\begin{center}
\small
\begin{tabular}{lcc}
\hline
name            & kernel dims        & output dims \\
 &  ($d\times h \times w$) & ($C\times D\times H\times W$) \\
\hline
conv1           & $3\times3\times3$ & $64\times8\times300\times400$ \\
max-pool1       & $1\times2\times2$ & $64\times8\times150\times200$ \\
conv2           & $3\times3\times3$ & $128\times8\times150\times200$ \\
max-pool2       & $2\times2\times2$ & $128\times4\times75\times100$ \\
conv3a          & $3\times3\times3$ & $256\times4\times75\times100$ \\
conv3b          & $3\times3\times3$ & $256\times4\times75\times100$ \\
max-pool3       & $2\times2\times2$ & $256\times2\times38\times50$ \\
conv4a          & $3\times3\times3$ & $512\times2\times38\times50$ \\
conv4b          & $3\times3\times3$ & $512\times2\times38\times50$ \\
max-pool4       & $2\times2\times2$ & $512\times1\times19\times25$ \\
conv5a          & $3\times3\times3$ & $512\times1\times19\times25$ \\
conv5b          & $3\times3\times3$ & $512\times1\times19\times25$ \\
\hline
toi-pool2*      & --                & $128\times8\times8\times8$ \\
toi-pool5       & --                & $512\times1\times4\times4$ \\
1x1 conv        & --                & $8192$ \\
\hline
fc6             & --                & $4096$ \\
fc7             & --                & $4096$ \\
\hline
\end{tabular}
\caption{Network architecture of T-CNN. We refer kernel with shape $d\times h \times w$ where $d$ is the kernel depth, $h$ and $w$ are height and width. Output matrix with shape $C\times D\times H\times W$ where $C$ is number of channels, $D$ is the number of frames, $H$ and $W$ are the height and width of frame. toi-pool2 only exists in TPN.}
\label{tab:architecture}
\end{center}
\end{table}

To better capture the spatio-temporal information in video, we exploit 3D CNN for action proposal generation and action recognition. One advantage of 3D CNN over 2D CNN is that it captures motion information by applying convolution in both time and space. Since 3D convolution and 3D max pooling are utilized in our approach, not only in the spatial dimension but also in the temporal dimension, the size of video clip is reduced while distinguishable information is concentrated. As demonstrated in \cite{c3d}, the temporal pooling is important for recognition task since it better models the spatio-temporal information of video and reduces some background noise. However, the temporal order information is lost. That means if we arbitrarily change the order of the frames in a video clip, the resulting 3D max pooled feature cube will be the same. This is problematic in action {\em detection}, since it relies on the feature cube to get bounding boxes for the original frames. To this end, we incorporate temporal skip pooling to retain temporal order information residing in the original frames. More details are provided in the next section.

\begin{figure}[!htb]
\centering
\includegraphics[width=0.65\linewidth]{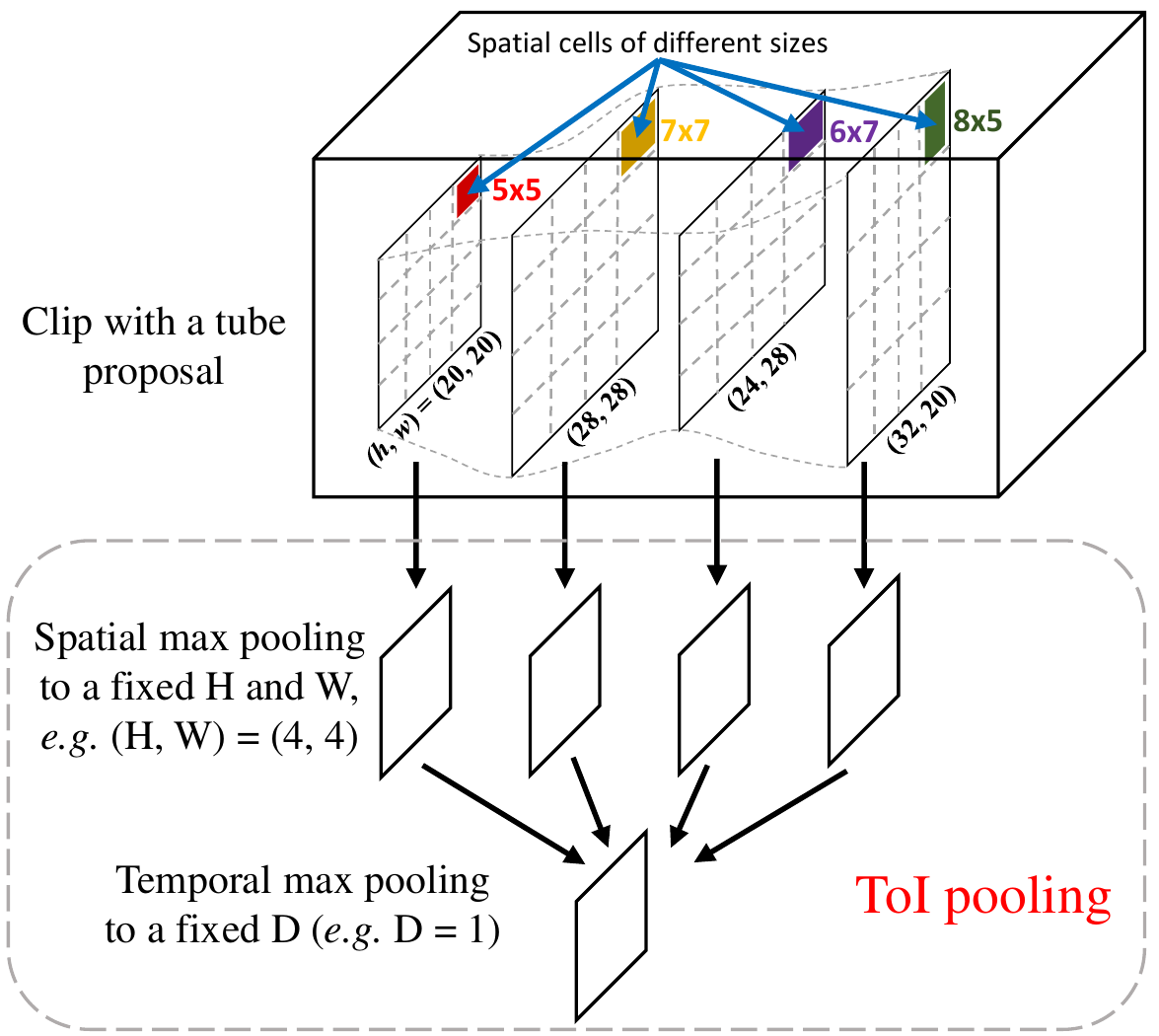}
\caption{Tube of interest pooling.}
\label{fig:toi_pool}
\end{figure}

Since a video is processed clip by clip, action tube proposals with various spatial and temporal sizes are generated for different clips. These clip proposals need to be linked into a  tube proposal sequence, which  is used for action label prediction and localization. To produce a fixed length feature vector, we propose a new pooling layer -- {\bf Tube-of-Interest} (ToI) pooling layer. The ToI pooling layer is a 3D generalization of Region-of-Interest (RoI) pooling layer of R-CNN. The classic max pooling layer defines the kernel size, stride and padding which determines the shape of the output. In contrast, for RoI pooling layer, the output shape is fixed first, then the kernel size and stride are determined accordingly. Compared to RoI pooling which takes 2D feature map and 2D regions as input, ToI pooling deals with feature cube and  3D tubes. Denote the size of a feature cube as $d\times h\times w$, where $d$, $h$ and $w$ respectively represent the depth, height and width of the feature cube. A ToI in the feature cube is defined by a $d$-by-$4$ matrix, which is composed of $d$ boxes distributed in all the frames. The boxes are defined by a four-tuple $(x_{1}^{i}, y_{1}^{i}, x_{2}^{i}, y_{2}^{i})$ that specifies the top-left and bottom-right corners in the $i$-th feature map. Since the $d$ bounding boxes may have different sizes, aspect ratios and positions, in order to apply spatio-temporal pooling, pooling in spatial and  temporal domains are performed separately. First, the $h\times w$ feature maps are divided  into $H\times W$ bins, where each bin corresponds to a cell with size of approximately $h/H\times w/W$. In each cell, max pooling is applied to select the maximum value. Second, the spatially pooled $d$ feature maps are temporally divided into $D$ bins. Similar to the first step, $d/D$ adjacent feature maps are grouped together to perform the standard temporal max pooling.  As a result the fixed output size of ToI pooling layer is $D \times H \times W$. A graphical illustration of ToI pooling is presented in Figure \ref{fig:toi_pool}.

Back-propagation of ToI pooling layer routes the derivatives from output back to the input. Assume $x_{i}$ is the $i$-th activation to the ToI pooling layer, and $y_{j}$ is the $j$-th output. Then the partial derivative of the loss function ($L$) with respect to each input variable $x_i$ can be expressed as:
\begin{align}
\frac{\partial L}{\partial x_{i}} = \sum_{j}[i = f(j)]\frac{\partial L}{\partial y_{j}}.
\end{align}
Each pooling output $y_j$ has a corresponding input position $i$. We use a function $f(\cdot)$ to represent the argmax selection. Thus, the gradient from the next layer $\partial L / \partial y_{j}$ is passed back to only that neuron which achieved the max $\partial L / \partial x_{i}$. Since one input may correspond to multiple outputs, the partial derivatives are the accumulation of multiple sources.

\section{T-CNN Pipeline}
\label{sec:approach}

\begin{figure*}[t]
	\centering
	\includegraphics[width=0.8\linewidth]{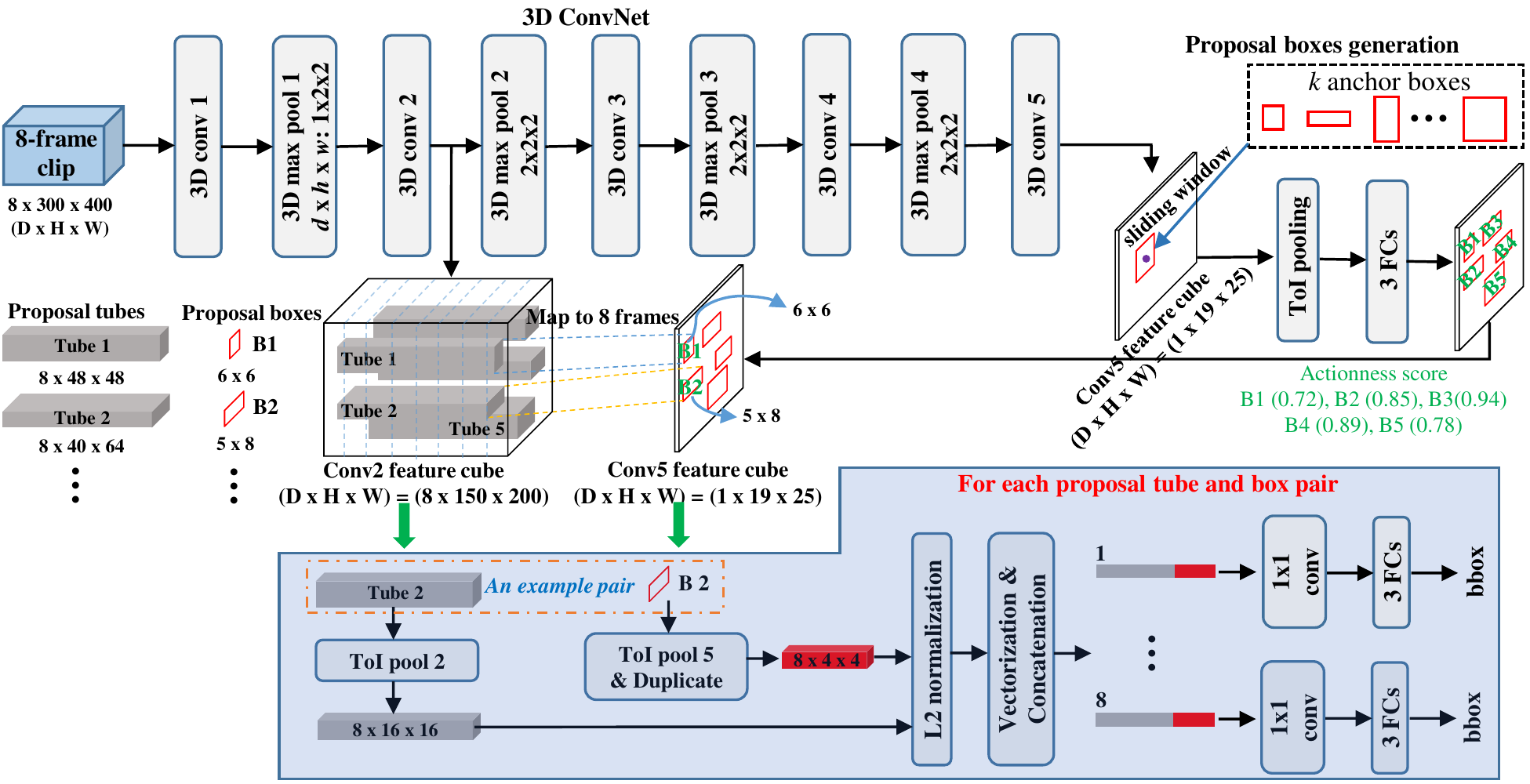}
	\caption{Tube proposal network. }
	\label{fig:TPN}
\end{figure*}

As shown in Figure \ref{fig:Teaser}, our T-CNN is an end-to-end deep learning framework that takes video clips as input. The core component is the Tube Proposal Network (TPN) (see Figure \ref{fig:TPN}) to produce tube proposals for each clip. Linked tube proposal sequence represents spatio-temporal action detection in the video and is also used for action recognition.

\subsection{Tube Proposal Network}
\label{sub-sec: tpn}
For a 8-frame video clip, 3D convolution and 3D pooling are used to extract spatio-temporal feature cube.
In 3D ConvNet, convolution and pooling are performed spatio-temporally. Therefore, the temporal information of the input video is preserved. Our 3D ConvNet consists of seven 3D convolution layers and four 3D max-pooling layers. We denote the kernel shape of 3D convolution/pooling by $d\times h\times w$, where $d, h, w$ are depth, height and width, respectively. In all convolution layers, the kernel sizes are $3\times 3\times 3$, padding and stride remain as $1$. The numbers of filters are 64, 128 and 256 respectively in the first $3$ convolution layers and $512$ in the remaining convolution layers.  The kernel size is set to $1\times 2\times 2$ for the first 3D max-pooling layer, and $2\times 2\times 2$ for the remaining 3D max-pooling layers. The details of network architecture are presented in Table \ref{tab:architecture}.  We use the C3D model \cite{c3d} as the pre-trained model and  fine tune it on each dataset in our experiments.

After conv5, the temporal size is reduced to 1 frame (\ie feature cube with depth $D = 1$). \textit{In the feature tube, each frame/slice consists of a number of channels specified in Table \ref{tab:architecture}. Here, we drop the number of channels for ease of explanation.} Following faster R-CNN, we generate bounding box proposals based on the conv5 feature cube.

\textbf{Anchor bounding boxes selection.} In faster R-CNN, the bounding box dimensions are hand picked, \ie 9 anchor boxes with 3 scales and 3 aspect ratios. We can directly adopt the same anchor boxes in our T-CNN framework. However, it has been shown in \cite{yolo9000} that if we choose better priors as initializations for the network, it will help the network learn better for predicting good detections.
Therefore, instead of choosing hand-picked anchor boxes, we apply k-means
clustering on the training set bounding boxes to learn $12$ anchor boxes (\ie clustering centroids). This data driven anchor box selection approach is adaptive to different datasets.

Each bounding box is associated with an ``actionness" score, which measures the probability that the content in the box corresponds to a valid action. We assign a binary class label (of being an action or not) to each bounding box. Bounding boxes with actionness scores smaller than a threshold are discarded. In the training phase, the bounding box which has an IoU overlap higher than 0.7 with any ground-truth box or has the highest Intersection-over-Union (IoU) overlap with a ground-truth box (the later condition is considered in case the former condition may find no positive sample) is considered as a positive bounding box proposal.

\textbf{Temporal skip pooling.} Bounding box proposals generated from conv5 feature tube can be used for frame-level action detection by bounding box regression. However, due to temporal concentration (8 frames to 1 frame) of temporal max pooling, the temporal order  of the original 8 frames is lost. Therefore, we use temporal skip pooling to inject the temporal order  for frame-level detection. Specifically, we map each positive bounding box generated from conv5 feature tube to conv2 feature tube which has 8 feature frames/slices. Since these 8 feature slices correspond to the original 8 frames in the video clip, the temporal order information is preserved. As a result, if there are 5 bounding boxes in conv5 feature tube for example, 5 scaled bounding boxes are mapped in each conv2 feature slice at the corresponding locations. This creates 5 tube proposals as illustrated in Figure \ref{fig:TPN}, which are paired with the corresponding 5 bounding box proposals for frame-level action detection. To form a fixed feature shape, ToI pooling is applied to the variable size tube proposals as well as the bounding box proposals. Since a tube proposal covers 8 frames, the ToI pooled bounding box is duplicated 8 times to form a tube. We then L2 normalize the paired two tubes and perform vectorization. For each frame, features are concatenated.
Since we use the C3D model \cite{c3d} as the pre-trained model, we connect a 1x1 convolution to match the input dimension of fc6.
Three fully-connected layers process each descriptor and produce the output: displacement of height, width and center coordinate of each bounding box (``bbox") in each frame. Finally, a set of refined tube proposals are generated as an output from the TPN representing spatio-temporal action localization of the input video clip.

\subsection{Linking Tube Proposals}
\label{subsec: link}

\begin{figure}[h]
\centering
\includegraphics[width=0.95\columnwidth]{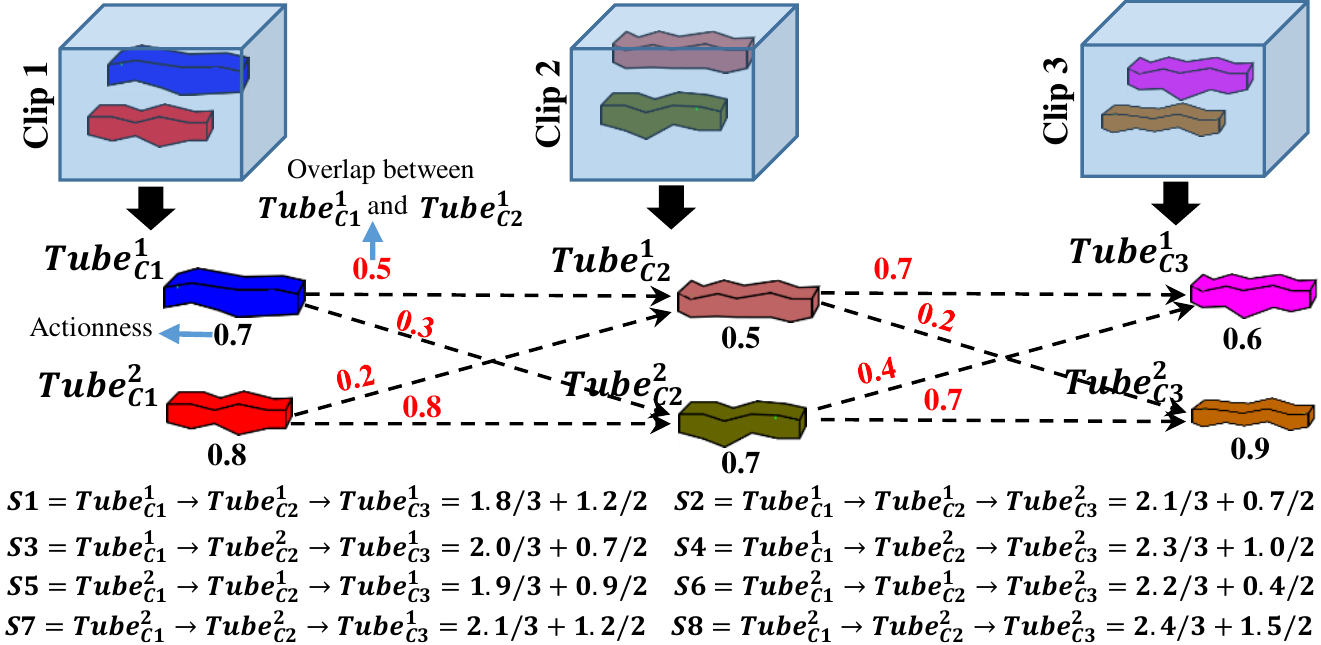}
\caption{An example of linking tube proposals in each video clips using network flow. In this example, there are three video clips and each has two tube proposals, resulting in 8 video proposals. Each video proposal has a score, \eg $S1, S2, ..., S8$, which is computed according to Eq. (2).}
\label{fig:link}
\end{figure}

We obtain  a set of tube proposals for each video clip after the TPN. We then link these tube proposals to form a proposal sequence for spatio-temporal action localization of the entire video.
Each tube proposal from different clips can be linked in a tube proposal sequence (\ie video tube proposal) for action detection. However, not all combinations of tube proposals can correctly capture the complete action. For example, a tube proposal in one clip may contain the action and a tube proposal in the following clip may only capture the background.  Intuitively, the content within the selected tube proposals should capture an action and connected tube proposals in any two consecutive clips should have a large temporal overlap. Therefore, two criteria are considered when linking tube proposals: actionness and overlap scores. Each video proposal is then assigned a score defined as follows:
\begin{equation}
S=\frac{1}{m}\sum_{i=1}^m Actionness_i+\frac{1}{m-1}\sum_{j=1}^{m-1} Overlap_{j,j+1}
\end{equation}
where $Actionness_i$ denotes the actionness score of the tube proposal from the $i$-th clip, $Overlap_{j,j+1}$ measures the overlap between the linked two proposals respectively from the $j$-th and $(j+1)$-th clips, and $m$ is the total number of video clips. As shown in Figure \ref{fig:TPN}, each bounding box proposal from conv5 feature tube is associated with an actionness score. The actionness scores are inherited by the corresponding tube proposals. The overlap between two tube proposals is calculated based on the IoU (Intersection Over Union) of the last frame of the $j$-th tube proposal and the first frame of the $(j+1)$-th tube proposal. The first term of $S$ computes the average actionness score of all tube proposals in a video proposal and the second term computes the average overlap between the tube proposals in every two consecutive video clips. Therefore, we ensure the linked tube proposals can encapsulate the action and at the same time have temporal consistency. An example of linking tube proposals and computing scores is illustrated in Figure \ref{fig:link}.
We choose a number of linked proposal sequences with highest scores in a video (see more details in Sec.~\ref{subsec:basic_settings}).

\subsection{Action Detection}
\label{subsec: recognition}
After linking tube proposals, we get a set of linked tube proposal sequences, which represent potential action instances. The next step is to classify these linked tube proposal sequences. The tube proposals in the linked sequences may have different sizes. In order to extract a fixed length feature vector from each of the linked proposal sequence, our proposed ToI pooling is utilized. Then the ToI pooling layer is followed by two fully-connected layers and a drop-out layer. The dimension of the last fully-connected layer is $N+1$ ($N$ action classes and $1$ background class). 

\section{Experiments}
\label{sec:experiments}

To verify the effectiveness of the proposed T-CNN for action detection,
we evaluate T-CNN on three trimmed video datasets including UCF-Sports \cite{rodriguez2008action},
J-HMDB \cite{Jhuang:ICCV:2013}, UCF-101 \cite{THUMOS13} and one un-trimmed video dataset -- THUMOS'14 \cite{THUMOS14}.

\subsection{Implementation Details}
\label{subsec:basic_settings}
We implement our method based on the Caffe toolbox \cite{jia2014caffe}. The TPN and recognition network share weights in their common layers. Due to memory limitation, in training phase, each video is divided into overlapping $8$-frame clips with resolution $300\times400$ and temporal stride $1$. When training the TPN network, each anchor box is assigned a binary label. Either the anchor box which has the highest IoU overlap with a ground-truth box, or an anchor box that has an IoU overlap higher than 0.7 with any ground-truth box is assigned a positive label, the rest are assigned negative label. In each iteration, $4$ clips are fed into the network. Since the number of background boxes is much more than that of action boxes, to well model the action, we randomly select some of the negative boxes to balance the number of positive and negative samples in a batch. For recognition network training, we choose 40 linked proposal sequences with highest scores in a video as Tubes of Interest.

 Our model is trained in an alternative manner. First, \textbf {Initialize TPN} based on the pre-trained model in \cite{c3d}, then using the generated proposals to \textbf{initialize recognition networks}. Next, the weights tuned by recognition network are used to \textbf{update TPN}. Finally, the tuned weights and proposals from TPN are used for \textbf{finalizing recognition network}. For all the networks for UCF-Sports and J-HMDB, the learning rate is initialized as $10^{-3}$ and decreased to $10^{-4}$ after $30$k batches. Training terminates after $50$k batches. For UCF-101 and THUMOS'14, the learning rate is initialized as $10^{-3}$ and decreased to $10^{-4}$ after $60$k batches. Training terminates after $100$k batches.

During testing, each video is divided into non-overlapping $8$-frame clips. If the number of frames in video cannot be divided by $8$, we pad zeros after the last frame to make it dividable. $40$ tube proposals with highest actionness confidence through TPN are chosen for the linking process. Non-maximum suppression (NMS) is applied to linked proposals to get the final action detection results.

\subsection{Datasets and Experimental Results}
\label{subsec:results}

\textbf{UCF-Sports.} This dataset contains 150 short action videos of 10 different sport classes. Videos are trimmed to the action and bounding boxes annotations are provided for all frames.
We follow the standard training and test split defined in \cite{lan2011discriminative}.

\begin{figure*}[!thb]
\centering
\includegraphics[width=0.99\linewidth]{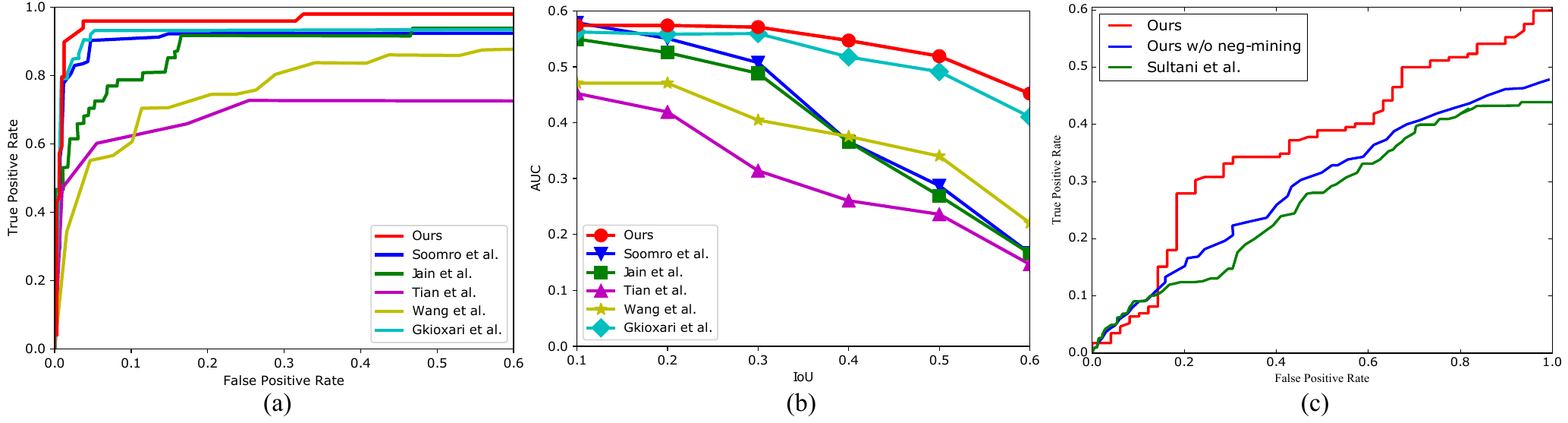}
\caption{The ROC and AUC curves for UCF-Sports Dataset \cite{rodriguez2008action} are shown in (a) and (b), respectively. The results are shown for Jain \etal \cite{jain2014action} (green), Tian \etal \cite{tian2013spatiotemporal} (purple), Soomro \etal \cite{soomro2015action} (blue), Wang \etal \cite{wang2014video} (yellow), Gkioxari \etal \cite{gkioxari2015finding} (cyan) and Proposed Method (red). (c) shows the mean ROC curves for four actions of THUMOS'14. The results are shown for Sultani \etal \cite{Sultani_2016_CVPR} (green), proposed method (red) and proposed method without negative mining (blue).}
\label{fig:curve_result}
\end{figure*}

We use the usual IoU criterion and generate ROC curve in Figure \ref{fig:curve_result}(a) when overlap criterion equals to $\alpha = 0.2$. Figure \ref{fig:curve_result}(b) illustrates AUC (Area-Under-Curve) measured with different overlap criterion. In direct comparison, our T-CNN clearly outperforms all the competing methods shown in the plot. We are unable to directly compare the detection accuracy against Peng \etal \cite{2stream-rcnn_peng:hal-01349107} in the plot, since they do not provide the ROC and AUC curves. As shown in Table \ref{tab:per-class}, the frame level mAP of our approach outperforms theirs in 8 actions out of 10. Moreover, by using the same metric, the video mAP of our approach reaches 95.2 ($\alpha = 0.2$ and $0.5$), while they report 94.8 ($\alpha = 0.2$) and 94.7 ($\alpha = 0.5$).

\begin{table*}[!ht]
\begin{center}
\small
\begin{tabular}{lccccccccccc}
\hline
                    & Diving    & Golf  & Kicking   & Lifting   & Riding    & Run   & SkateB.     & Swing     & SwingB.   & Walk  & mAP \\
\hline
Gkioxari \etal \cite{gkioxari2015finding}     & 75.8      & 69.3  & 54.6      & 99.1      & 89.6      & 54.9  & 29.8              & 88.7      & 74.5      & 44.7  & 68.1 \\
Weinzaepfel \etal \cite{weinzaepfel2015learning}  & 60.71     & 77.55 & 65.26     & {\bf 100.00}    & 99.53     & 52.60 & 47.14             & {\bf 88.88}     & 62.86     & 64.44 & 71.9 \\
Peng \etal \cite{peng2016multi}         & {\bf 96.12}     & 80.47 & 73.78     & 99.17     & 97.56     & 82.37 & 57.43             & 83.64     & 98.54     & 75.99 & 84.51 \\
Ours                & 84.38     &  {\bf 90.79} & {\bf 86.48 }    & 99.77      & {\bf 100.00}    & {\bf 83.65} & {\bf 68.72 }            & 65.75     & {\bf 99.62}     & {\bf 87.79} & {\bf 86.7} \\
\hline
\end{tabular}
\end{center}
\caption{mAP for each class of UCF-Sports. The IoU threshold $\alpha$ for frame m-AP is fixed to $0.5$.}
\label{tab:per-class}
\end{table*}

\textbf{J-HMDB.}  This dataset consists of 928 videos with 21 different actions. All the video clips are well trimmed. There are three train-test splits and the evaluation is done on the average results over the three splits. The experiment results comparison is shown in Table \ref{tab:jhmdb}. We report our results with 3 metrics: frame-mAP, the average precision of detection at frame level as in \cite{gkioxari2015finding}; video-mAP, the average precision at video level as in \cite{gkioxari2015finding} with IoU threshold $\alpha=0.2$ and $\alpha=0.5$. It is evident that our T-CNN consistently outperforms the state-of-the-art approaches in terms of all three evaluation metrics.

\begin{table}[!ht]
\begin{center}
\small
\begin{tabular}{lccc}
\hline
                                                    & f.-mAP        & v.-mAP        & v.-mAP  \\
                                                    & ($\alpha=0.5$)& ($\alpha=0.2$)& ($\alpha=0.5$) \\
\hline
Gkioxari \etal \cite{gkioxari2015finding}           & 36.2          & --            & 53.3 \\
Weinzaepfel \etal \cite{weinzaepfel2015learning}    & 45.8          & 63.1          & 60.7 \\
Peng \etal \cite{peng2016multi}                     & 58.5          & 74.3          & 73.1 \\
Ours w/o skip pooling                               & 47.9          & 66.9          & 58.6 \\
Ours                                                & {\bf 61.3 }   & {\bf 78.4 }   & {\bf 76.9}\\
\hline
\end{tabular}
\end{center}
\caption{Comparison to the state-of-the-art on J-HMDB. The IoU threshold $\alpha$ for frame m-AP is fixed to 0.5.}
\label{tab:jhmdb}
\end{table}

\textbf{UCF101.} This dataset with 101 actions is commonly used for action recognition. For action detection task, a subset of $24$ action classes and $3,207$ videos have spatio-temporal annotations. Similar to other methods, we perform the experiments on the first train/test split only. We report our results in Table \ref{tab:ucf101} with 3 metrics: frame-mAP, video-mAP ($\alpha=0.2$) and video-mAP ($\alpha=0.5$). Our approach again yields the best performance. Moreover, we also report the action recognition results of T-CNN on the above three datasets in Table \ref{tab:rec}.

\begin{table}[!ht]
\begin{center}
\footnotesize
\begin{tabular}{lccccc}
\hline
                    & f.-mAP & \multicolumn{4}{c}{video-mAP} \\
IoU th.             &           & $0.05$   & $0.1$    & $0.2$    & $0.3$ \\
\hline
Weinzaepfel \etal \cite{weinzaepfel2015learning}   & 35.84     & 54.3              & {\bf51.7 }              & 46.8              & 37.8 \\
Peng \etal \cite{peng2016multi}         & 39.63     & 54.5              & 50.4              & 42.3              & 32.7 \\
Ours                & {\bf 41.37 }    & {\bf 54.7 }             & 51.3             & {\bf 47.1}              & {\bf 39.2} \\
\hline
\end{tabular}
\end{center}
\caption{Comparison to the state-of-the-art on UCF-101 (24 actions). The IoU threshold $\alpha$ for frame m-AP is fixed to 0.5.}
\label{tab:ucf101}
\end{table}


\textbf{THUMOS'14.}  To further validate the effectiveness of our proposed T-CNN approach for action detection, we evaluate it using the untrimmed videos from the THUMOS'14 dataset \cite{THUMOS14}. The THUMOS'14 spatio-temporal localization task consists of $4$ classes of sports actions: BaseballPitch, golfSwing, TennisSwing and ThrowDiscus. There are about $20$ videos per class and each video contains $500$ to $3,000$ frames. The videos are divided into validation set and test set, but only video in the test set have spatial annotations provided by \cite{Sultani_2016_CVPR}. Therefore, we use samples corresponding to those 4 actions in UCF-101 with spatial annotations to train our model.

In untrimmed videos, there often exist other unrelated actions besides the action of interests. For example, ``walking" and ``picking up a golf ball" are considered as unrelated actions when detecting ``GolfSwing" in video. We denote clips which have positive ground truth annotation as positive clips, and the other clips as negative clips (\ie clips contain only unrelated actions). If we randomly select negative samples for training, the number of boxes on unrelated actions is much smaller than that of background boxes (\ie boxes capturing only image background). Thus the trained model will have no capability to distinguish action of interest and unrelated actions.

To this end, we introduce a so called \textbf{negative sample mining} process. Specifically, when initializing the TPN, we only use positive clips. Then we apply the model on the whole training video (both positive clips and negative clips). Most false positives in negative clips should include unrelated actions to help our model learn the correlation between action of interest and unrelated actions. Therefore we select boxes in negative clips with highest scores as \textbf{hard negatives} because low scores probably infer image background. In updating TPN procedure, we choose 32 boxes which have IoU with any ground truth greater than 0.7 as positive samples and randomly pick another 16 samples as negative. We also select 16 samples from hard negative pool as negative. Therefore, we efficiently train a model, which is able to distinguish not only action of interest from background, but also action of interest from unrelated actions.

The mean ROC curves of different methods on THUMOS'14 action detection are plotted in Figure \ref{fig:curve_result}(c). Our method without negative mining performs better than the baseline method Sultani \etal \cite{Sultani_2016_CVPR}. Additionally, with negative mining, the performance is further boosted.


\begin{figure*}[!t]
\centering
\includegraphics[width=0.99\linewidth]{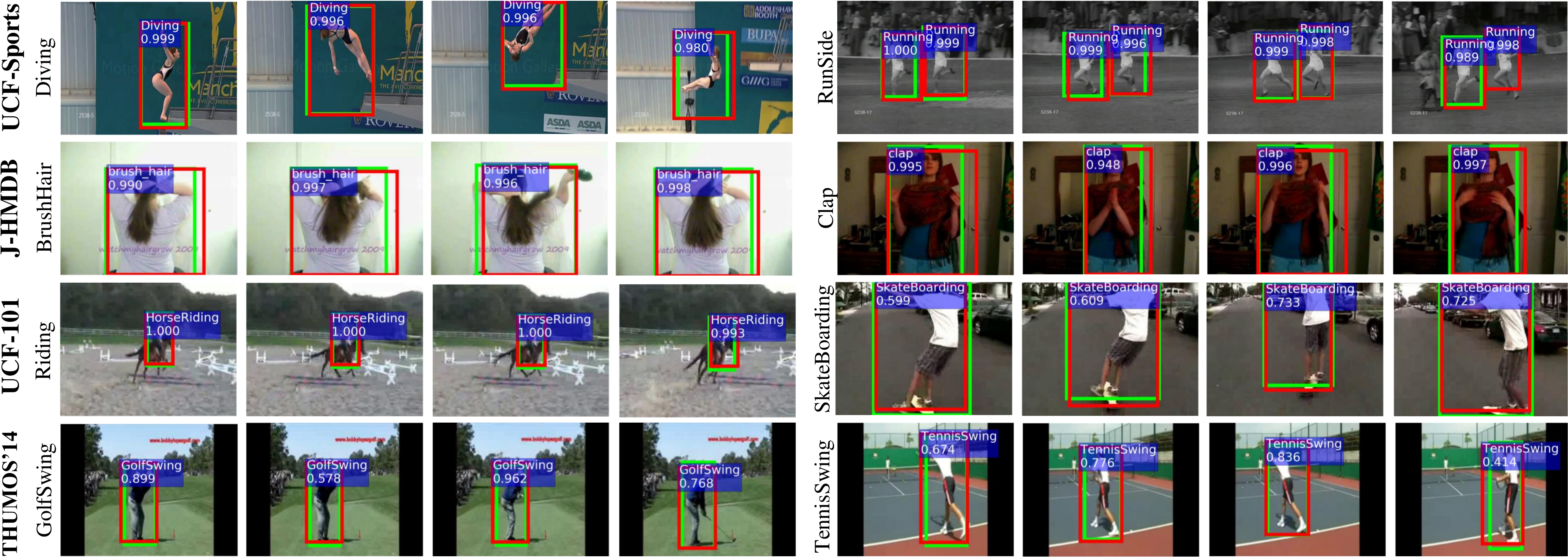}
\caption{Action detection results by T-CNN on UCF-Sports, JHMDB, UCF-101 and THUMOS'14. Red boxes indicate the detections in the corresponding frames, and green boxes denote ground truth. The predicted label is overlaid. }
\label{fig:results}
\end{figure*}

For qualitative results, we shows examples of detected action tubes in videos from UCF-Sports, JHMDB, UCF-101 (24 actions) and THUMOS'14 datasets (see Figure \ref{fig:results}). Each block corresponds to a different video that is selected from the test set. We show the highest scoring action tube for each video.

\begin{table}[!tb]
\begin{center}
\small
\begin{tabular}{lcccc}
\hline
                        & Accuracy (\%)  \\
\hline
UCF-Sports              & 95.7  \\
J-HMDB                  & 67.2       \\
UCF-101 (24 actions)    & 94.4   \\
\hline

\end{tabular}
\caption{Action recognition results of our T-CNN approach on the four datasets.}
\label{tab:rec}
\end{center}
\end{table}

\section{Discussion}
\label{sec:discussion}

\subsection{ToI Pooling}
\label{subsec:toipool}
To evaluate the effectiveness of ToI pooling, we compare action recognition performance on UCF-101 dataset (101 actions) using C3D \cite{c3d} and our approach. For the C3D network, we use C3D pre-train model from \cite{c3d} and fine tune the weights on UCF-101 dataset. In the C3D fine tuning process, a video is divided into $16$ frames clips first. Then the C3D network is fed by clips and outputs a feature vector for each clip. Finally, a SVM classifier is trained to predict the labels of all clips, and the video label is determined by the predictions of all clips belonging to the video. Compared to the original C3D network, we use the ToI pooling layer to replace the 5-th 3d-max-pooling layer in C3D pipeline.
Similar to C3D network, our approach takes clips from a video as input. The ToI pooling layer takes the whole clip as tube of interest and the pooled depth is set to 1. As a result, each video will output one feature vector. Therefore, it is an end-to-end deep learning based video recognition approach. Video level accuracy is used as the metric. The results are shown in Table \ref{tab:toi_eval}.
For a direct comparison, we only use the result from deep network without fusion with other features. Our approach shows a $5.2\%$ accuracy improvement compared to the original C3D. Our ToI pooling based pipeline optimizes the weights for the whole video directly, while C3D performs clip-based optimization. Therefore, our approach can better capture the spatio-temporal information of the entire video. Furthermore, our ToI pooling can be combined with other deep learning based pipelines, such as two-stream CNN \cite{2stream_cnn_simonyan_2014two}.

\begin{table}[!h]
\small
\begin{center}
\begin{tabular}{lcc}
\hline
                & C3D \cite{c3d}    & Ours \\
\hline
Accuracy (\%)   & 82.3              & 87.5 \\
\hline

\end{tabular}
\caption{Video action recognition results on UCF-101.}
\label{tab:toi_eval}
\end{center}
\end{table}

\subsection{Temporal Skip Connection}
Since we use overlapping clips with temporal stride of $1$ in training, a particular frame is included in multiple training clips at different temporal positions. The actual temporal information of that particular frame is lost if we only use the conv5 feature cube to infer action bounding boxes. Especially when the action happens periodically (\ie SwingBench), it always fails to locate a phase of spinning. On the contrary, by combining conv5 with conv2 through temporal skip pooling, temporal order is preserved to localize actions more accurately. To verify the effectiveness of temporal skip pooling in our proposed TPN, we conduct an experiment using our method without skip connection. In other words, we perform bounding box regression to estimate bounding boxes in 8 frames simultaneously using only the conv5 feature cube. As shown in Table \ref{tab:jhmdb}, without skip connection, the performance decreases a lot, demonstrating the advantage of skip connection for extracting temporal order information and detailed motion in original frames.


\subsection{Computational Cost}
We carry out our experiments on a workstation with one GPU (Nvidia GTX Titan X). For a 40-frames video, it takes 1.1 seconds to generate tube proposals, 0.03 seconds to link tube proposals in a video and 0.9 seconds to predict action label.

\section{Conclusion}
\label{sec:conclusion}
In this paper we propose an end-to-end Tube Convolutional Neural Network (T-CNN) for action detection in videos. It exploits 3D convolutional network to extract effective spatio-temporal features and perform action localization and recognition in a unified framework. Coarse proposal boxes are densely sampled based on the 3D convolutional feature cube and linked for action recognition and localization.  Extensive experiments on several benchmark datasets demonstrate the strength of T-CNN for spatio-temporal localizing actions, even in untrimmed videos.

\vspace{0.75em}
\noindent \small{\textbf{Acknowledgement.} The project was supported by Award No. 2015-R2-CX-K025, awarded by the National Institute of Justice, Office of Justice Programs, U.S. Department of Justice. The opinions, findings, and conclusions or recommendations expressed in this publication are those of the author(s) and do not necessarily reflect those of the Department of Justice.}


\begin{thebibliography}{10}\itemsep=-1pt

\bibitem{dpm}
P.~Felzenszwalb, D.~McAllester, and D.~Ramanan.
\newblock A discriminatively trained, multiscale, deformable part model.
\newblock In {\em Computer Vision and Pattern Recognition, 2008. CVPR 2008.
  IEEE Conference on}, pages 1--8. IEEE, 2008.

\bibitem{gaidon2013temporal}
A.~Gaidon, Z.~Harchaoui, and C.~Schmid.
\newblock Temporal localization of actions with actoms.
\newblock {\em IEEE transactions on pattern analysis and machine intelligence},
  35(11):2782--2795, 2013.

\bibitem{fast_rcnn_Girshick_2015_ICCV}
R.~Girshick.
\newblock Fast r-cnn.
\newblock In {\em The IEEE International Conference on Computer Vision (ICCV)},
  December 2015.

\bibitem{rcnn_Girshick_2014_CVPR}
R.~Girshick, J.~Donahue, T.~Darrell, and J.~Malik.
\newblock Rich feature hierarchies for accurate object detection and semantic
  segmentation.
\newblock In {\em The IEEE Conference on Computer Vision and Pattern
  Recognition (CVPR)}, June 2014.

\bibitem{gkioxari2015finding}
G.~Gkioxari and J.~Malik.
\newblock Finding action tubes.
\newblock In {\em Proceedings of the IEEE Conference on Computer Vision and
  Pattern Recognition}, pages 759--768, 2015.

\bibitem{jain2014action}
M.~Jain, J.~Van~Gemert, H.~J{\'e}gou, P.~Bouthemy, and C.~G. Snoek.
\newblock Action localization with tubelets from motion.
\newblock In {\em Proceedings of the IEEE Conference on Computer Vision and
  Pattern Recognition}, pages 740--747, 2014.

\bibitem{jain201515}
M.~Jain, J.~C. van Gemert, and C.~G. Snoek.
\newblock What do 15,000 object categories tell us about classifying and
  localizing actions?
\newblock In {\em 2015 IEEE Conference on Computer Vision and Pattern
  Recognition (CVPR)}, pages 46--55. IEEE, 2015.

\bibitem{Jhuang:ICCV:2013}
H.~Jhuang, J.~Gall, S.~Zuffi, C.~Schmid, and M.~J. Black.
\newblock Towards understanding action recognition.
\newblock In {\em International Conf. on Computer Vision (ICCV)}, pages
  3192--3199, Dec. 2013.

\bibitem{ji20133d}
S.~Ji, W.~Xu, M.~Yang, and K.~Yu.
\newblock 3d convolutional neural networks for human action recognition.
\newblock {\em IEEE transactions on pattern analysis and machine intelligence},
  35(1):221--231, 2013.

\bibitem{jia2014caffe}
Y.~Jia, E.~Shelhamer, J.~Donahue, S.~Karayev, J.~Long, R.~Girshick,
  S.~Guadarrama, and T.~Darrell.
\newblock Caffe: Convolutional architecture for fast feature embedding.
\newblock {\em arXiv preprint arXiv:1408.5093}, 2014.

\bibitem{THUMOS13}
Y.-G. Jiang, J.~Liu, A.~Roshan~Zamir, I.~Laptev, M.~Piccardi, M.~Shah, and
  R.~Sukthankar.
\newblock {THUMOS} challenge: Action recognition with a large number of
  classes.
\newblock \url{/ICCV13-Action-Workshop/}, 2013.

\bibitem{THUMOS14}
Y.-G. Jiang, J.~Liu, A.~Roshan~Zamir, G.~Toderici, I.~Laptev, M.~Shah, and
  R.~Sukthankar.
\newblock {THUMOS} challenge: Action recognition with a large number of
  classes.
\newblock \url{http://crcv.ucf.edu/THUMOS14/}, 2014.

\bibitem{yolo9000}
R.~Joseph and F.~Ali.
\newblock Yolo9000: Better, faster, stronger.
\newblock In {\em arXiv:1612.08242v1}, 2016.

\bibitem{karpathy2014large}
A.~Karpathy, G.~Toderici, S.~Shetty, T.~Leung, R.~Sukthankar, and L.~Fei-Fei.
\newblock Large-scale video classification with convolutional neural networks.
\newblock In {\em Proceedings of the IEEE conference on Computer Vision and
  Pattern Recognition}, pages 1725--1732, 2014.

\bibitem{ke2007event}
Y.~Ke, R.~Sukthankar, and M.~Hebert.
\newblock Event detection in crowded videos.
\newblock In {\em 2007 IEEE 11th International Conference on Computer Vision},
  pages 1--8. IEEE, 2007.

\bibitem{lan2011discriminative}
T.~Lan, Y.~Wang, and G.~Mori.
\newblock Discriminative figure-centric models for joint action localization
  and recognition.
\newblock In {\em Computer Vision (ICCV), 2011 IEEE International Conference
  on}, pages 2003--2010. IEEE, 2011.

\bibitem{lecun2015deep}
Y.~LeCun, Y.~Bengio, and G.~Hinton.
\newblock Deep learning.
\newblock {\em Nature}, 521(7553):436--444, 2015.

\bibitem{action_survey}
F.~Negin and F.~Bremond.
\newblock Human action recognition in videos: A survey.
\newblock 2016.

\bibitem{peng2016multi}
X.~Peng and C.~Schmid.
\newblock Multi-region two-stream r-cnn for action detection.
\newblock In {\em European Conference on Computer Vision}, pages 744--759.
  Springer, 2016.

\bibitem{2stream-rcnn_peng:hal-01349107}
X.~Peng and C.~Schmid.
\newblock {Multi-region two-stream R-CNN for action detection}.
\newblock In {\em {ECCV 2016 - European Conference on Computer Vision}},
  Amsterdam, Netherlands, Oct. 2016.

\bibitem{faster_rcnn}
S.~Ren, K.~He, R.~Girshick, and J.~Sun.
\newblock Faster r-cnn: Towards real-time object detection with region proposal
  networks.
\newblock In C.~Cortes, N.~D. Lawrence, D.~D. Lee, M.~Sugiyama, and R.~Garnett,
  editors, {\em Advances in Neural Information Processing Systems 28}, pages
  91--99. Curran Associates, Inc., 2015.

\bibitem{rodriguez2008action}
M.~Rodriguez, A.~Javed, and M.~Shah.
\newblock Action mach: a spatio-temporal maximum average correlation height
  filter for action recognition.
\newblock In {\em Computer Vision and Pattern Recognition, 2008. CVPR 2008.
  IEEE Conference on}, pages 1--8. IEEE, 2008.

\bibitem{2stream_cnn_simonyan_2014two}
K.~Simonyan and A.~Zisserman.
\newblock Two-stream convolutional networks for action recognition in videos.
\newblock In {\em Advances in Neural Information Processing Systems}, pages
  568--576, 2014.

\bibitem{soomro2015action}
K.~Soomro, H.~Idrees, and M.~Shah.
\newblock Action localization in videos through context walk.
\newblock In {\em Proceedings of the IEEE International Conference on Computer
  Vision}, pages 3280--3288, 2015.

\bibitem{Sultani_2016_CVPR}
W.~Sultani and M.~Shah.
\newblock What if we do not have multiple videos of the same action? -- video
  action localization using web images.
\newblock In {\em The IEEE Conference on Computer Vision and Pattern
  Recognition (CVPR)}, June 2016.

\bibitem{sun2015human}
L.~Sun, K.~Jia, D.-Y. Yeung, and B.~E. Shi.
\newblock Human action recognition using factorized spatio-temporal
  convolutional networks.
\newblock In {\em Proceedings of the IEEE International Conference on Computer
  Vision}, pages 4597--4605, 2015.

\bibitem{tian2013spatiotemporal}
Y.~Tian, R.~Sukthankar, and M.~Shah.
\newblock Spatiotemporal deformable part models for action detection.
\newblock In {\em Proceedings of the IEEE Conference on Computer Vision and
  Pattern Recognition}, pages 2642--2649, 2013.

\bibitem{c3d}
D.~Tran, L.~Bourdev, R.~Fergus, L.~Torresani, and M.~Paluri.
\newblock Learning spatiotemporal features with 3d convolutional networks.
\newblock In {\em 2015 IEEE International Conference on Computer Vision
  (ICCV)}, pages 4489--4497. IEEE, 2015.

\bibitem{wang2014video}
L.~Wang, Y.~Qiao, and X.~Tang.
\newblock Video action detection with relational dynamic-poselets.
\newblock In {\em European Conference on Computer Vision}, pages 565--580.
  Springer, 2014.

\bibitem{weinzaepfel2015learning}
P.~Weinzaepfel, Z.~Harchaoui, and C.~Schmid.
\newblock Learning to track for spatio-temporal action localization.
\newblock In {\em Proceedings of the IEEE International Conference on Computer
  Vision}, pages 3164--3172, 2015.

\bibitem{lstm_ng}
J.~Yue-Hei~Ng, M.~Hausknecht, S.~Vijayanarasimhan, O.~Vinyals, R.~Monga, and
  G.~Toderici.
\newblock Beyond short snippets: Deep networks for video classification.
\newblock In {\em Proceedings of the IEEE Conference on Computer Vision and
  Pattern Recognition}, pages 4694--4702, 2015.

\end{thebibliography}

{\small
\bibliographystyle{ieee}

}

\section{Appendix}
In this supplemental material, we provide more discussions on our proposed T-CNN framework for action detection.
We also include some video clips showing the detection results of our T-CNN approach.

\subsection{Analysis of Temporal Skip Pooling}
In our paper, we already compared the performance of T-CNN with and without temporal skip pooling on J-HMDB dataset (Table 3). Here we show more results on UCF-Sports dataset to demonstrate the power of temporal skip pooling.

All the experiments use the same parameters. The only difference is which feature cubes are connected with the feature cube from conv5, denoted as $C_5$.

\begin{table}[!hpbt]
\begin{center}
\begin{tabular}{l|c|cc}
\hline
Feature From        & f. mAP        & f. mAP        & v. mAP \\
                    & $\alpha = 0.5$& $\alpha = 0.2$& $\alpha = 0.5$ \\
\hline
$C_5$               & 74.9          & 91.6          & 77.9 \\
$C_5 + C_1$         & 81.2          & 94.5          & 92.1 \\
$C_5 + C_2$         & \textbf{86.7} & \textbf{95.2} & \textbf{94.8} \\
$C_5 + C_3$         & 85.8          & \textbf{95.2} & 91.7 \\
$C_5 + C_4$         & 77.6          & 91.3          & 81.2 \\
\hline
\end{tabular}
\end{center}
\end{table}

As shown in the table, $C_5$ only has the lowest performance since it is not able to distinguish the differences between frames. $C_5 + C_1$ perform much better than $C_5$ only. Since conv1 output maintains the temporal structure (8 frames) and preserves detailed motion information. However, the first convolution layer only captures some low level properties, which may not be useful. $C_5 + C_2$, which we reported in our submission, has the best performance. On one hand, conv2 maintains the temporal structure as well. On the other hand, it captures more high level information than conv1, which may be directly related to some particular classes. Since conv3 feature cube has a reduced temporal dimension of $4$, it preserves less temporal information. Therefore the performance of $C_5 + C_3$ is slightly worse than $C_5 + C_2$. For the same reason, $C_5 + C_4$ perform better than $C_5$, but worse than $C_5 + C3$. In summary, the spatio-temporal localization model needs descriptive as well as distinguishing capabilities. The output of latter convolutional layers are more distinguishable, but due to the temporal collapse, they have less descriptive property. When we concatenate conv5 with conv2, which preserves much more  spatial and temporal information, the better performance is achieved. It is also worth noting that more convolution layers can be concatenated \eg $C_5 + C_2 + C_3$ to further improve the performance. However, it will increase the computational burden as well as memory cost. As a trade off between performance and computational \& memory efficiency, we only concatenate two convolutional feature cubes in our experiments.


\subsection{More Action Detection Results}
Besides the detection results presented in the paper (Figure 6), we include some video clips showing the detection results. The videos can be viewed at \url{https://www.dropbox.com/sh/6bakdb9s88u8zif/AAB3SxJkvvtdQm0JxtYVLkMma?dl=0}

The first 4 videos are from different datasets. The green boxes represent the groundtruth annotation and red ones are our detection results. All the videos are encoded by DivX contained in an AVI.

The $5$-th video is downloaded from YouTube, which does not appear in any existing dataset. We use the model trained on UCF-Sports and detect the runners. This video is much more complex than UCF-Sports. The background has crowded people with motions \eg \textit{waving hands} and \textit{waving flags}, and the camera motion is more severe. Here, we show some key frames of the clip in Figure A. As can be seen, our T-CNN is able to successfully detect the action ``running" from both runners given complicated background and noise disturbance (\eg camera motion and background actions). This demonstrates the generalization ability of our approach to action detection in videos with complex actions.

\begin{figure}[!h]
\includegraphics[width=0.95\columnwidth]{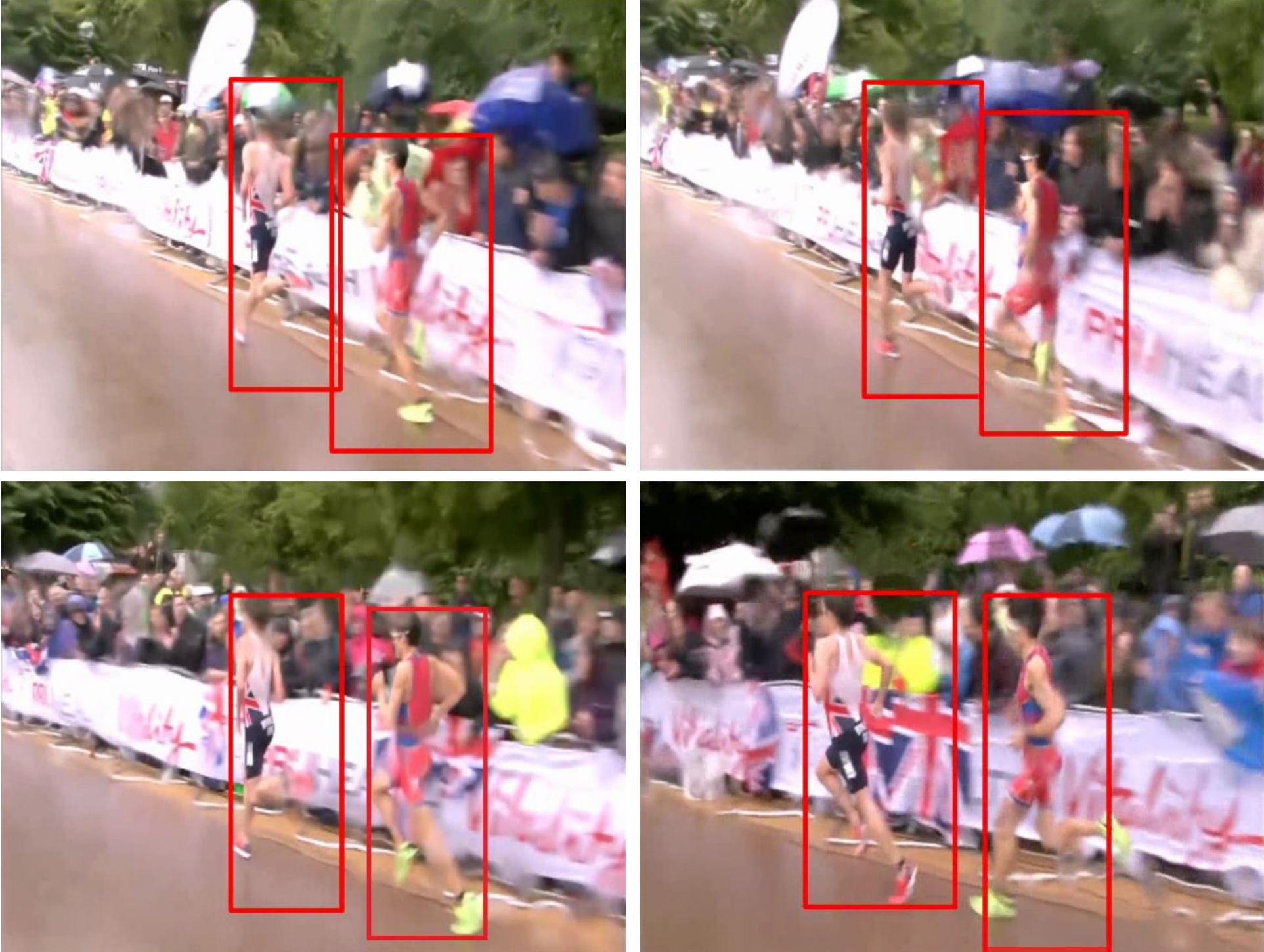}
\caption*{Figure A. Example frames of action detection results on a YouTube video ``running".}
\end{figure}

The $6$-th and $7$-th videos are another two such videos downloaded from YouTube, corresponding to actions ``horse riding" and ``diving", respectively. Example action detection results of our T-CNN for these two videos are presented in Figure B and Figure C.

\begin{figure}[!h]
\includegraphics[width=0.95\columnwidth]{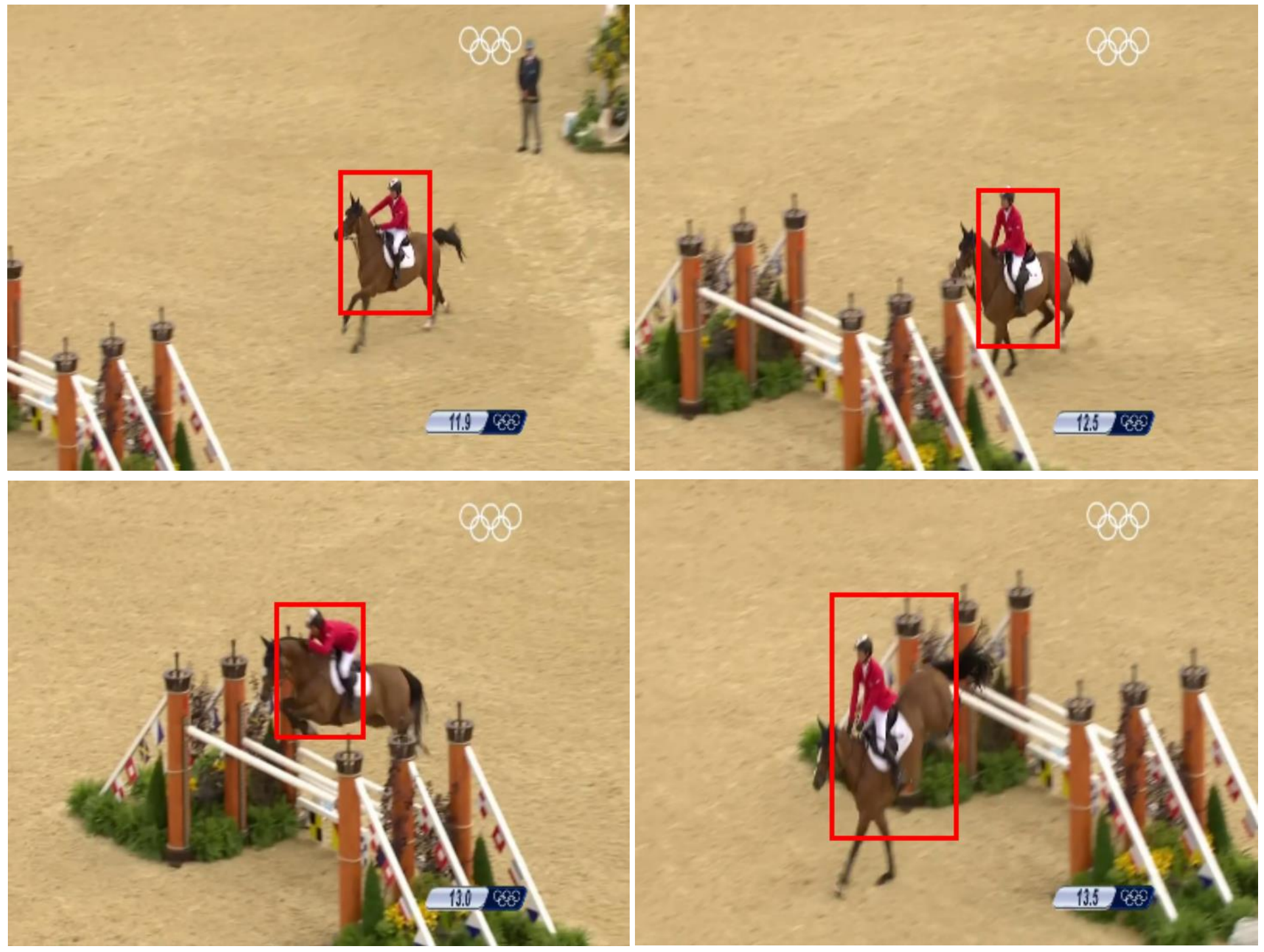}
\caption*{Figure B. Example frames of action detection results on a YouTube video ``horse riding".}
\end{figure}

\begin{figure}[!h]
\includegraphics[width=0.95\columnwidth]{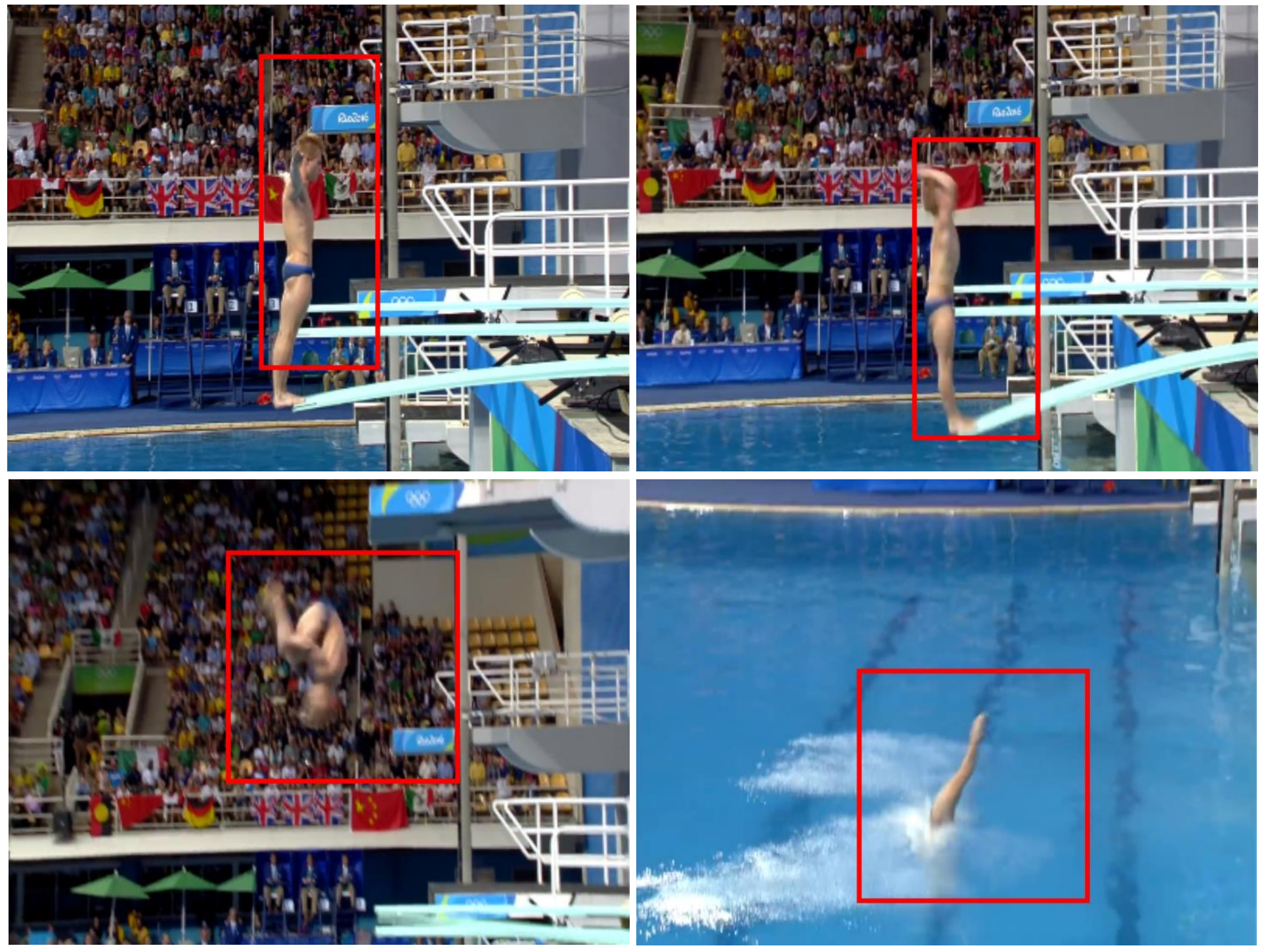}
\caption*{Figure C. Example frames of action detection results on a YouTube video ``diving".}
\end{figure}

\end{document}